\pdfoutput=1
\documentclass{article}

\PassOptionsToPackage{numbers,compress}{natbib}

\usepackage[main, final]{nips_requirements/neurips_2026}

\usepackage[utf8]{inputenc}
\usepackage[T1]{fontenc}
\usepackage{url}
\usepackage{booktabs}
\usepackage{amsfonts}
\usepackage{amsmath}
\usepackage{amssymb}
\usepackage{nicefrac}
\usepackage{microtype}
\usepackage[table]{xcolor}
\usepackage{graphicx}
\usepackage{wrapfig}
\usepackage{multirow}
\usepackage{makecell}
\usepackage{threeparttable}
\usepackage{array}
\usepackage{subcaption}
\usepackage{adjustbox}
\usepackage{tabularx}
\usepackage{pdflscape}
\usepackage{enumitem}
\usepackage{float}
\usepackage{listings}
\usepackage{amsthm}
\usepackage{bm}
\usepackage{algorithm}
\usepackage{algpseudocode}

\usepackage{hyperref}

\graphicspath{{./}{./figs/}}

\title{
Rethinking the Readout: Unlocking \\
Video Backbones for AI-Generated Video Detection}
\date{}

\author{
  \normalfont\normalsize
  Manni Cui\textsuperscript{1,*}\hspace{0.85em}%
  Ziheng Qin\textsuperscript{2,*}\hspace{0.85em}%
  ZiAn Wang\textsuperscript{3,*}\hspace{0.85em}%
  Ruiqi Liu\textsuperscript{2,*}\hspace{0.85em}%
  Dianyuan Zou\textsuperscript{1}\hspace{0.85em}%
  Jianglan Wei\textsuperscript{1} \\
  Han Zhou\textsuperscript{1}\hspace{0.85em}%
  Yu Liu\textsuperscript{1}\hspace{0.85em}%
  Jingrui Xu\textsuperscript{1}\hspace{0.85em}%
  Wenhao Wang\textsuperscript{4,\textdagger}\hspace{0.85em}%
  Zhenyu Zhang\textsuperscript{1,\textdagger} \\[12pt]
  \makebox[\dimexpr\textwidth-2\tabcolsep\relax][c]{%
    \textsuperscript{1}Huazhong University of Science and Technology \quad
    \textsuperscript{2}Institute of Automation, Chinese Academy of Sciences} \\
  \makebox[\dimexpr\textwidth-2\tabcolsep\relax][c]{%
    \textsuperscript{3}Jilin University \quad
    \textsuperscript{4}Vast Intelligence Lab} \\[3pt]
  \textsuperscript{*}Equal contribution \quad
  \textsuperscript{\textdagger}Corresponding author
}

\begin{document}

\maketitle
\vspace{-10pt}
\begin{figure}[H]
      \centering
      \includegraphics[width=\linewidth]{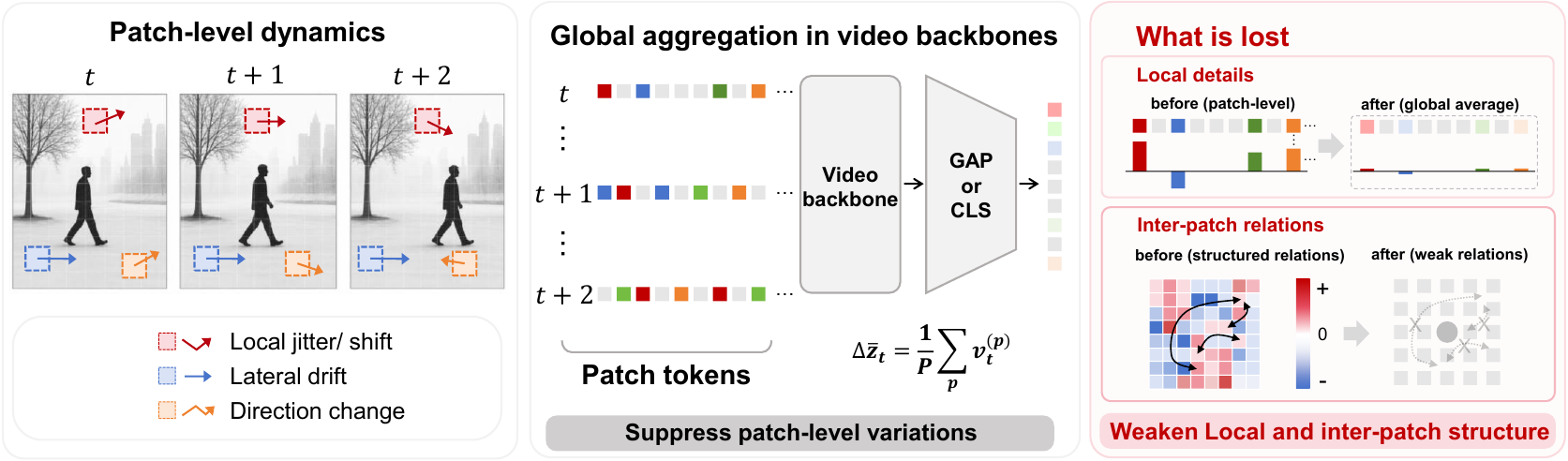}
      \vspace{5pt} 
      \caption{\textbf{The readout is the bottleneck for AIGV detection.} Left: AIGV detection relies on patch level temporal cues. Middle: Each colored cell represents a patch token, with color depth indicating activation strength. Standard video backbones aggregate these tokens into a single global vector via GAP or CLS, which suppresses patch level variations. Right: This collapse weakens local details and erases structured inter patch relations, leaving the bottleneck in the readout rather than in the backbone representation.}
      \label{fig:aigv_teaser}
\end{figure}
\vspace{-14pt}
\begin{abstract}
AI-generated videos (AIGVs) typically contain subtle temporal artifacts that arise from inter-frame inconsistencies rather than within individual frames. A detector that captures such artifacts should therefore benefit from video pretrained backbones over image only ones. In practice, however, video backbones with standard global readouts often fail to outperform strong image pretrained probes on AIGV benchmarks. We attribute this gap to excessive spatiotemporal aggregation in the readout. Video pretrained backbones tend to compress each frame into a single global descriptor. This compression suppresses local patch level temporal dynamics and discards inter patch relations, which are precisely the cues that AIGV detection most reliably depends on. Based on this, we propose Velocity Gated Patch Velocity Profiling (V-PVP), a lightweight readout that replaces only the aggregation layer with two parallel streams over the patch velocity field, adding only about $0.5$M trainable parameters. V-PVP serves as a general plug-and-play module that consistently improves performance across diverse video backbones under both end-to-end fine-tuning and linear probing settings. Our method reaches \textbf{95.28} AUC on AIGVDBench while keeping the backbone fully frozen. The results show that simply replacing the aggregation layer reactivates the temporal potential of frozen video backbones, restoring their advantage on AIGV detection. Code is available at \url{https://github.com/manic-cui/Rethinking-the-readout}.
\end{abstract}

\newpage
\section{Introduction}
\label{sec:intro}

Recent video generation models\citep{OpenAI2024Sora, hong2022cogvideo, yang2024cogvideox, kong2024hunyuanvideo, zheng2024open, wan2025wan} produce synthetic videos that are increasingly difficult to distinguish from real ones. Their misuse raises pressing concerns about misinformation, which makes it crucial to identify discriminative signals that separate real videos from generated ones. Many of these signals are temporal in nature, manifesting as inter-frame inconsistencies such as flicker, jitter, drift, or motion that is inconsistent across regions \citep{chen2024demamba, bai2024ai, liu2024turns, hayun2026training}.

A natural choice for exploiting such temporal signals is video backbones pretrained on large scale video datasets, since they are designed to model temporal dynamics across frames. Recent benchmarks, however, suggest the opposite. Several detectors built on image pretrained encoders \citep{tan2024rethinking, chen2025dual, yan2024orthogonal} have been shown to surpass video backbones with their default readouts on standard AIGV detection benchmarks \citep{ma2026your}. This suggests that the temporal advantage of video backbones is not being fully realized, and that the bottleneck likely lies elsewhere in the pipeline.

We trace this gap to a specific bottleneck and conduct a diagnostic study in Section~\ref{sec:motivation}, identifying excessive spatiotemporal aggregation in the readout as the cause. Video pretraining objectives, from supervised action recognition to self-supervised masked video modeling, all push the backbone toward a stable global representation, with patch-level variations treated as noise. This pipeline suppresses local patch level dynamics and discards inter patch relations, both of which carry the most reliable cues for distinguishing real from generated videos \citep{zhang2025physics, kim2025beyond}. The bottleneck therefore lies in the readout rather than in the backbone representation.

Building on this diagnosis, we propose Velocity Gated Patch Velocity Profiling (V-PVP), a lightweight readout that keeps the backbone fully frozen and replaces only the aggregation layer. V-PVP addresses the two losses identified above with two parallel streams over the patch velocity field. One stream uses velocity gated patch attention to focus on patches with notable temporal variations, which would otherwise be smoothed out by uniform aggregation. The other stream aggregates patch velocities to preserve their per channel magnitudes, which would otherwise be diluted by linear averaging. The new readout introduces only about $0.5$M trainable parameters, roughly $0.57\%$ of the backbone.

Our contributions are as follows:
\begin{itemize}[leftmargin=*,itemsep=2pt]
    \item We identify the readout as an important bottleneck of video backbones in AIGV detection, and provide quantitative evidence that standard aggregation discards two complementary forms of patch level evidence on which AIGV detection relies.
    \item We propose V-PVP, a readout that addresses the two losses with two parallel streams over the patch velocity field, unlocking the perceptual potential of video backbones for AIGV detection.
    \item Across multiple video backbones and benchmarks, V-PVP consistently surpasses same backbone readouts to a large extent. With VideoMAE as the backbone, our method reaches \textbf{95.28} AUC on AIGVDBench while keeping the backbone fully frozen.
\end{itemize}

\section{Related Work}
\label{sec:related}

\subsection{AI-Generated Video Detection}

As video generation has rapidly advanced, AIGV detection has shifted from single-frame artifact analysis toward modeling temporal consistency. Existing methods broadly fall into three groups. End-to-end spatiotemporal detectors train or fine tune a discriminator on AIGV data, including AIGVDet, DeMamba, DIVID, and recent extensions \citep{bai2024ai, chen2024demamba, liu2024turns, chang2024matters, ma2025detecting, li2025skyra, jiang2025ivy}, but they often memorize the training generator's spatial fingerprint and generalize poorly to unseen generators \citep{tan2024rethinking, zhou2025breaking}. Methods built on physical or frequency priors inject inter frame frequency, motion coherence, or spectral artifacts as discriminative anchors \citep{kim2025beyond, zhang2025physics, chen2025gc, qian2020thinking}. Train free approaches use only real video statistics with representation priors and avoid seeing fake data, gaining robustness at the cost of in-distribution accuracy \citep{hayun2026training, jang2025cinemae}. In addition, several detectors built on frozen image encoders \citep{ojha2023towards, liu2026mirror, tan2024rethinking} extend to video by aggregating per frame scores and remain competitive against video-backbone counterparts. Across these lines, the readout layer is treated as a minor implementation detail, leaving the temporal advantage of video backbones largely unexploited.

\subsection{Video Backbones and Temporal Readouts}

Common video backbones such as TimeSformer \citep{bertasius2021space}, VideoMAE \citep{tong2022videomae}, UniFormerV2 \citep{li2022uniformerv2}, SlowFast \citep{feichtenhofer2019slowfast}, I3D \cite{carreira2017quo}, and XCLIP \citep{ni2022expanding} are pretrained on action recognition datasets such as Kinetics-400 \citep{kay2017kinetics}. Downstream detection pipelines typically reuse the action-recognition readout, collapsing patch tokens into a global CLS or GAP vector. This pattern has carried over to deepfake-domain ViTs \citep{coccomini2022combining, heo2021deepfake, wodajo2021deepfake, wang2023deep} with patch-level self-attention and pooling variants, none of which tie the aggregation to a generator-invariant physical quantity. Action recognition rewards stable dominant motion, while AIGV detection hinges on subtle local inconsistencies, so the standard readout is unlikely to be the right fit. To our knowledge, V-PVP is the first method to diagnose the readout as the bottleneck of video backbones in AIGV detection and to address it with a minimal-parameter replacement.

\begin{figure}[t]
    \centering
    \includegraphics[width=\linewidth]{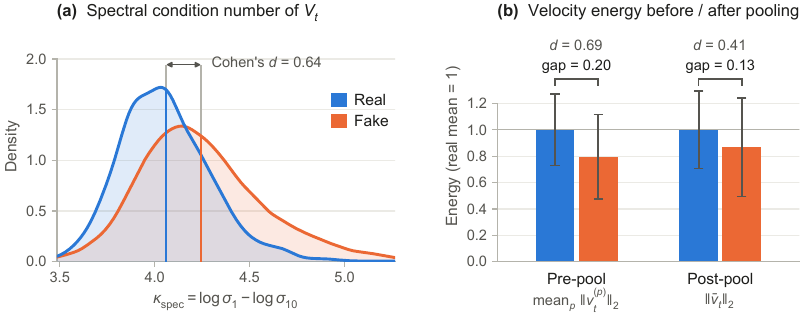}
    \caption{\textbf{The readout is the bottleneck.} Real vs fake statistics on all 63k AIGVDBench test videos. (a)~Loss 1. Spectral condition number of the patch velocity matrix $V_t \in \mathbb{R}^{P \times D}$, defined as $\kappa_{\text{spec}} = \log \sigma_1 - \log \sigma_{10}$; vertical lines mark the group means. Generated videos are systematically higher, with Cohen's $d \approx 0.64$. This signal is no longer measurable after aggregation. (b)~Loss 2. Per channel velocity energies, normalized so that the real video mean equals $1$; whiskers show $\pm 1$ SD. Pre pool, real and fake energies differ by $1.00$ vs $0.80$ with Cohen's $d \approx 0.69$. After pooling, the gap shrinks to $1.00$ vs $0.87$ with $d \approx 0.41$.}
    \label{fig:readout-bottleneck}
\end{figure}

\section{Motivation}
\label{sec:motivation}

A standard video backbone takes an input clip and outputs a tensor of patch tokens of shape $P \times T \times D$, where $P$ is the number of spatial patches per frame, $T$ is the temporal length, and $D$ is the channel dimension. These tokens carry rich spatiotemporal information, but downstream classification heads expect a single fixed length vector. The readout layer is what closes this gap. Common choices, such as global average pooling, a learnable CLS token, or fixed query attention pooling, differ in how the aggregation weights are computed but share the same operation, namely a joint spatiotemporal and semantic aggregation over patch tokens.

This kind of aggregation is the right design for tasks that rely on a stable global understanding, such as action recognition, where local frame to frame fluctuations and inter patch differences are noise to be averaged out. AIGV detection has the opposite need. The discriminative signal between real and generated videos lives precisely in those small frame to frame variations within each patch and in the relations among patches. Standard readouts smooth out the former and dissolve the latter in a single step, leaving the pooled vector blind to the very cues that AIGV detection most reliably depends on. We formalize this mismatch as two complementary losses, namely the loss of inter patch relations and the dilution of per channel velocity magnitudes.

To make this concrete, we focus on GAP as the simplest case and base the analysis on the patch velocity
\begin{equation}
    v_t^{(p)} = z_{t+1}^{(p)} - z_t^{(p)},
    \label{eq:velocity}
\end{equation}
where $z_t^{(p)} \in \mathbb{R}^{D}$ denotes the $p$-th patch token of frame $t$. Applying GAP across patches and then taking a temporal difference is equivalent to averaging the patch velocities,
\begin{equation}
    \Delta \bar z_t 
    = \bar z_{t+1} - \bar z_t 
    = \frac{1}{P}\sum_{p=1}^{P} z_{t+1}^{(p)} - \frac{1}{P}\sum_{p=1}^{P} z_t^{(p)}
    = \frac{1}{P}\sum_{p=1}^{P} \left( z_{t+1}^{(p)} - z_t^{(p)} \right)
    = \frac{1}{P}\sum_{p=1}^{P} v_t^{(p)}.
    \label{eq:gap-collapse}
\end{equation}
The patch axis collapses from $P \times D$ to $1 \times D$ in a single step, and an analogous collapse happens along the temporal axis. This collapse erases inter patch relations and cancels velocity components across channels, the two effects we analyze below.

\paragraph{Loss 1: inter patch relations are erased.}
Inter patch relations capture how different patches co-vary across frames. They live in the covariance structure across the patch dimension and become undefined once the patches are aggregated into a single vector. To probe whether real and generated videos differ in this structure, we measure the spectral condition number of the patch velocity matrix $V_t$ at each frame, which tests whether inter patch motion admits a low rank decomposition, that is, whether all patches move homogeneously. As shown in Figure~\ref{fig:readout-bottleneck}(a), generated videos consistently exhibit more homogeneous inter patch motion than real videos, reflecting their different generation processes. This pattern is a reliable cue for distinguishing real from fake, yet it lives in the patch dimension and disappears the moment GAP collapses it.
\paragraph{Loss 2: per channel velocity magnitudes are diluted.}
Patch velocity vectors typically point in different directions across patches, so when the linear average is taken, components with opposite signs cancel along each channel. By Jensen's inequality, the post-pool magnitude is bounded by the pre-pool average,
\begin{equation}
    \underbrace{\frac{1}{P}\sum_{p=1}^{P} \left\| v_t^{(p)} \right\|_2}_{\text{pre-pool magnitude}}
    \;\;\ge\;\; 
    \underbrace{\left\| \frac{1}{P}\sum_{p=1}^{P} v_t^{(p)} \right\|_2}_{\text{post-pool magnitude}}
    \;=\; \left\| \bar v_t \right\|_2,
    \label{eq:loss2-jensen}
\end{equation}
with equality only when all $v_t^{(p)}$ are colinear and same-signed. As shown in Figure~\ref{fig:readout-bottleneck}(b), pooling retains only a small fraction of the per channel energy, and the gap between real and generated videos shrinks substantially after pooling. Unlike Loss 1, the magnitude pattern still exists in the pooled vector, but common readouts can no longer reliably extract it.

The two losses share the same source. The standard readout is designed for joint spatiotemporal and semantic aggregation, which by its very design erases inter patch structure and dilutes channel level magnitudes. Recovering both requires intervening at the readout rather than within the backbone, as we describe in Section~\ref{sec:method}; a formal derivation of both losses and the corresponding repair operators is provided in Appendix~\ref{app:derivation}.

\begin{figure}[!t]
      \centering
      \includegraphics[width=\linewidth]{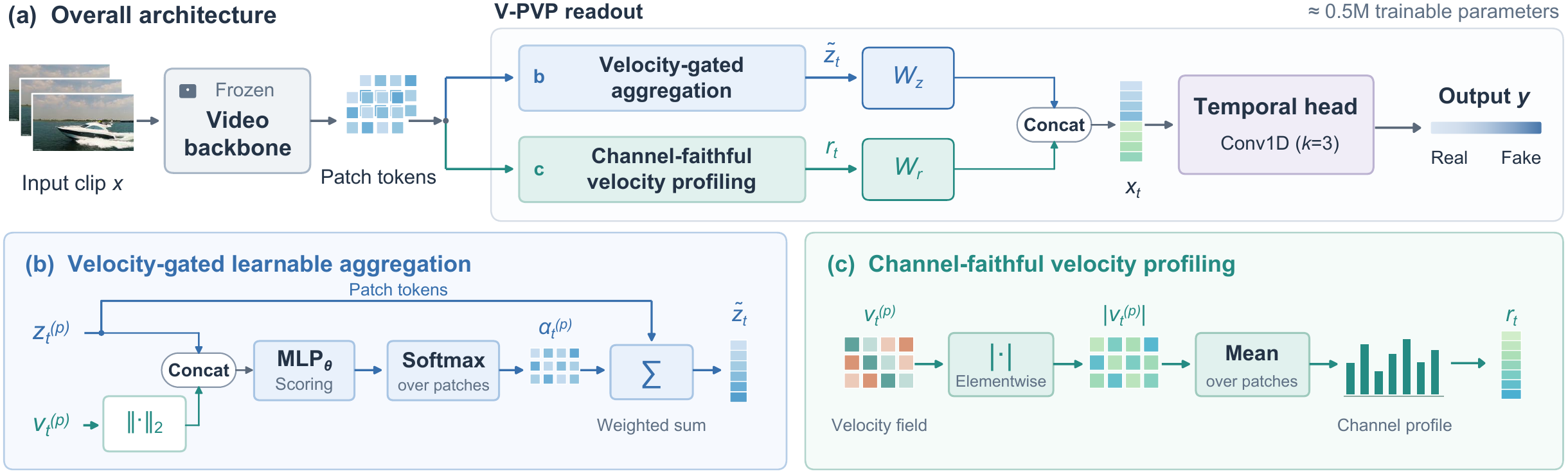}
      \caption{\textbf{Overall architecture of V-PVP.} (a) The video backbone stays frozen and only the readout (${\approx}0.5$M parameters) is trained: two streams over the patch velocity $v_t^{(p)} = z_{t+1}^{(p)} - z_t^{(p)}$ are projected by $W_z$ and $W_r$, concatenated into $x_t$, and read by a temporal head into a real/fake score. (b) An MLP scores each patch from $[z_t^{(p)}, \|v_t^{(p)}\|_2]$, and the softmax weights $\alpha_t^{(p)}$ pool the patch tokens into $\tilde z_t$. (c) The element-wise $|v_t^{(p)}|$, averaged over patches, gives $r_t$.}
      \label{fig:aigv_backbone}
\end{figure}

\section{Method}
\label{sec:method}

Figure~\ref{fig:aigv_backbone} shows the overall design of V-PVP. We keep the backbone fully frozen and replace only the readout with two parallel streams over the patch velocity field $\{v_t^{(p)}\}$. A velocity gated learnable aggregation $\tilde z_t$ (Section~\ref{sec:method:zt}) addresses Loss 1 by carrying inter patch heterogeneity into the pooled output, and a channel faithful velocity stream $r_t$ (Section~\ref{sec:method:rt}) addresses Loss 2 by preserving per channel magnitudes through an element wise nonlinearity. The two streams are projected independently, concatenated at each time step, and passed to a shallow temporal head (Section~\ref{sec:method:head}).

\subsection{Velocity-Gated Learnable Aggregation $\tilde z_t$}
\label{sec:method:zt}

Let $z_t^{(p)}$ be the $p$-th patch token of frame $t$ and $v_t^{(p)} = z_{t+1}^{(p)} - z_t^{(p)}$ the patch velocity. A small scoring network produces a per patch score on the fly, and a softmax converts scores into aggregation weights,
\begin{equation}
    s_t^{(p)}
    \;=\;
    \mathrm{MLP}_\theta\!\bigl([z_t^{(p)},\;\|v_t^{(p)}\|_2]\bigr),
    \quad
    \alpha_t^{(p)}
    \;=\;
    \mathrm{softmax}_p\!\bigl(s_t^{(p)}\bigr),
    \quad
    \tilde z_t
    \;=\;
    \sum_p \alpha_t^{(p)}\,z_t^{(p)}.
    \label{eq:vgap}
\end{equation}

Standard readouts collapse the patch dimension with weights that are fixed once training ends, applying the same weighting regardless of which patches are anomalous. $\tilde z_t$ differs in two respects. First, the learned $\alpha_t^{(p)}$ can vary by orders of magnitude across patches within a single frame, whereas the uniform $1/P$ of GAP dilutes every patch equally. Second, $\alpha_t^{(p)}$ is computed from the current video's $(z_t^{(p)},\,\|v_t^{(p)}\|_2)$, so the spatial distribution of weights itself encodes inter patch heterogeneity. This weight pattern takes different shapes across videos and propagates into the channel pattern of $\tilde z_t$ through the weighted sum. In effect, the $P$-dimensional weight vector $\alpha_t$ acts as a per video carrier of the inter patch structure that motivated Loss 1, allowing $\tilde z_t$ to retain a discriminative signature of this structure even after collapsing the patch axis.

The scoring network receives $\|v_t^{(p)}\|_2$ explicitly alongside $z_t^{(p)}$. Scoring on appearance alone reduces the aggregator to a semantic selector that overfits to the spatial fingerprint of the training generator (Table~\ref{tab:spatial}). Conditioning on velocity magnitude anchors the definition of ``anomalous'' to a generator invariant physical quantity and prevents fingerprint memorization.

\subsection{Channel-Faithful Velocity Stream $r_t$}
\label{sec:method:rt}

Although $\tilde z_t$ addresses Loss 1 by introducing input-adaptive patch weighting, it remains a signed linear aggregation over patch-level signals. Velocity components from different patches can still cancel on each channel when they have opposite signs, so a signed linear readout cannot guarantee faithful preservation of per-channel velocity magnitudes. Loss 2 therefore requires a nonlinear operation before spatial aggregation.

The $r_t$ stream avoids this by applying an element wise absolute value before the spatial sum,
\begin{equation}
    r_t
    \;=\;
    \frac{1}{P}\sum_p \bigl|v_t^{(p)}\bigr|.
    \label{eq:rt}
\end{equation}
Moving the nonlinearity inside the sum prevents opposite signed components from cancelling, so per channel velocity magnitude is faithfully preserved. The resulting vector encodes both how much anomalous activity exists and which channels it occupies.

We use the mean rather than the max because channel level anomalies in generated videos typically appear as weak signals distributed across many patches. The mean accumulates these diffuse contributions and is robust to isolated noise.

\begin{table}[t!]
    \centering
    \caption{Per-generator AUC on the \textbf{AIGVDBench} open-source split (20 generators). Trained to convergence (Section~\ref{sec:exp:setup}), unlike the 5-epoch ablations in Table~\ref{tab:ablation}. Best average in \textbf{bold}, second \underline{underlined}.}
    \label{tab:aigvd_open}
    \setlength{\tabcolsep}{1.8pt}
    \renewcommand{\arraystretch}{1.15}
    \resizebox{\textwidth}{!}{    \begin{tabular}{l|cccccc|cccccccccccc|cc|c}
        \toprule
        & \multicolumn{6}{c|}{\textbf{I2V}} & \multicolumn{12}{c|}{\textbf{T2V}} & \multicolumn{2}{c|}{\textbf{V2V}} & \\
        \cmidrule(lr){2-7} \cmidrule(lr){8-19} \cmidrule(lr){20-21}
        \textbf{Method}
        & \scriptsize\makecell{Easy\\Animate} & \scriptsize LTX & \scriptsize\makecell{Pyramid\\Flow} & \scriptsize SEINE & \scriptsize SVD & \scriptsize\makecell{Video\\Crafter}
        & \scriptsize\makecell{Acc\\Video} & \scriptsize\makecell{Animate\\Diff} & \scriptsize\makecell{Cogvideo\\x1.5} & \scriptsize\makecell{Easy\\Animate} & \scriptsize Hunyuan & \scriptsize IPOC & \scriptsize LTX & \scriptsize\makecell{Open\\Sora} & \scriptsize\makecell{Pyramid\\Flow} & \scriptsize\makecell{Rep\\Video} & \scriptsize\makecell{Video\\Crafter} & \scriptsize\makecell{Wan\\2.1}
        & \scriptsize\makecell{Cogvideo\\x1.5} & \scriptsize LTX
        & \textbf{AVG} \\
        \midrule
        \multicolumn{22}{c}{\textbf{AI-Generated Image Detection Models}} \\
        \midrule
        NPR \cite{tan2024rethinking} & 96.78 & 85.76 & 99.22 & 99.21 & 99.93 & 93.97 & 68.32 & 66.10 & 93.72 & 96.23 & 74.35 & 93.82 & 92.04 & 99.99 & 99.45 & 86.50 & 98.44 & 76.72 & 96.96 & 80.50 & 89.90 \\
        UnivFD \cite{ojha2023towards} & 80.18 & 82.77 & 86.83 & 94.29 & 96.39 & 80.01 & 82.74 & 93.29 & 94.77 & 90.77 & 83.27 & 97.49 & 93.30 & 99.72 & 97.23 & 90.36 & 98.03 & 84.16 & 93.47 & 81.56 & 90.03 \\
        Effort \cite{yan2024orthogonal} & 81.88 & 82.29 & 86.96 & 92.90 & 96.02 & 79.80 & 82.39 & 93.42 & 94.82 & 92.44 & 83.85 & 97.75 & 93.28 & 99.74 & 97.71 & 92.18 & 98.33 & 85.58 & 93.54 & 80.88 & 90.29 \\
        DDA \cite{chen2025dual} & 78.28 & 90.54 & 88.79 & 99.57 & 99.99 & 99.76 & 86.02 & 92.43 & 89.82 & 75.77 & 87.34 & 97.58 & 92.15 & 98.85 & 88.78 & 82.53 & 99.74 & 89.70 & 89.93 & 87.08 & 90.73 \\
        MIRROR \cite{liu2026mirror} & 55.03 & 43.59 & 30.93 & 98.05 & 93.31 & 99.86 & 69.73 & 90.45 & 51.15 & 73.20 & 67.15 & 74.45 & 45.66 & 43.27 & 49.63 & 66.62 & 99.87 & 70.06 & 54.63 & 54.76 & 66.57 \\
        \midrule
        \multicolumn{22}{c}{\textbf{AI-Generated Video Detection Models}} \\
        \midrule
        DeMamba \cite{chen2024demamba} & 98.50 & 96.25 & 98.75 & 99.25 & 97.50 & 95.50 & 66.50 & 85.25 & 95.75 & 98.50 & 73.00 & 96.25 & 97.00 & 98.25 & 97.75 & 90.50 & 95.75 & 88.75 & 98.50 & 95.25 & \underline{93.14} \\
        DeCoF \cite{ma2025detecting} & 90.84 & 90.55 & 95.54 & 97.33 & 98.98 & 63.10 & 88.25 & 89.24 & 98.16 & 91.87 & 86.33 & 99.33 & 97.61 & 99.76 & 99.76 & 97.22 & 89.25 & 83.53 & 97.40 & 89.86 & 92.20 \\
        ReSTrav \cite{interno2025ai} & 60.17 & 68.40 & 64.52 & 55.02 & 54.97 & 73.73 & 80.25 & 88.00 & 61.57 & 66.13 & 79.32 & 68.67 & 60.82 & 79.60 & 67.65 & 63.85 & 80.38 & 60.98 & 78.88 & 72.55 & 69.27 \\
        NSG-VD \cite{zhang2025physics} & 50.49 & 54.79 & 52.87 & 52.58 & 50.66 & 53.91 & 57.23 & 51.98 & 52.24 & 50.02 & 48.24 & 52.26 & 54.06 & 62.62 & 51.43 & 52.03 & 47.04 & 48.12 & 54.44 & 53.97 & 52.55 \\
        \midrule
        \multicolumn{22}{c}{\textbf{Video Classification Models}} \\
        \midrule
        I3D \cite{carreira2017quo} & 95.10 & 91.51 & 90.11 & 84.35 & 90.65 & 90.75 & 96.06 & 95.10 & 91.81 & 97.14 & 95.78 & 93.61 & 94.08 & 99.65 & 95.49 & 87.93 & 98.48 & 84.59 & 96.24 & 88.51 & 92.85 \\
        SlowFast \cite{feichtenhofer2019slowfast} & 93.47 & 94.45 & 95.42 & 94.41 & 98.60 & 91.50 & 91.11 & 88.24 & 93.27 & 96.53 & 91.59 & 91.97 & 97.71 & 97.64 & 96.71 & 88.18 & 97.24 & 73.08 & 91.34 & 91.61 & 92.70 \\
        TimeSformer \cite{bertasius2021space} & 78.86 & 78.38 & 80.94 & 77.16 & 84.74 & 84.26 & 87.15 & 91.60 & 87.16 & 89.31 & 86.77 & 88.38 & 85.84 & 96.57 & 91.03 & 80.82 & 94.77 & 78.33 & 88.01 & 78.20 & 85.41 \\
        UniFormerV2 \cite{li2022uniformerv2} & 76.92 & 68.96 & 74.77 & 73.47 & 76.82 & 67.75 & 71.21 & 73.98 & 75.76 & 81.35 & 72.37 & 80.81 & 73.22 & 95.34 & 86.86 & 74.75 & 84.27 & 78.87 & 73.92 & 67.77 & 76.46 \\
        VideoMAE \cite{tong2022videomae} & 84.21 & 84.15 & 87.11 & 92.81 & 88.24 & 94.52 & 91.64 & 94.28 & 81.67 & 88.04 & 88.82 & 87.11 & 86.78 & 98.74 & 93.13 & 80.93 & 97.62 & 76.86 & 87.76 & 82.45 & 88.34 \\
        \midrule
        \rowcolor[RGB]{210,225,250}
        \textbf{V-PVP (Ours)} & 94.18 & 94.05 & 94.68 & 97.48 & 94.53 & 97.19 & 97.75 & 96.84 & 93.82 & 94.82 & 95.90 & 96.11 & 95.82 & 99.69 & 96.40 & 90.17 & 99.02 & 87.67 & 97.45 & 91.96 & \textbf{95.28} \\
        \bottomrule
    \end{tabular}}
\end{table}
\subsection{Two-Stream Fusion and Sequence Head}
\label{sec:method:head}

At each frame pair, $\tilde z_t$ and $r_t$ each produce a $D$-dimensional vector with very different statistics. $\tilde z_t$ inherits the near zero mean signed distribution of post LayerNorm features, while $r_t$ is strictly non-negative and typically an order of magnitude smaller. A shared projection cannot rescale both. We apply two independent linear maps to $H/2$ each and concatenate along the channel axis at each time step,
\begin{equation}
    u_t = W_z\,\tilde z_t,
    \qquad
    w_t = W_r\,r_t,
    \qquad
    x_t = [u_t;\, w_t].
    \label{eq:fusion}
\end{equation}
Splitting the channel budget keeps the two signals on a common scale before fusion, and bounds the readout at the $0.5$M parameter level.

\begin{figure}[!b]
    \centering
    \includegraphics[width=\linewidth]{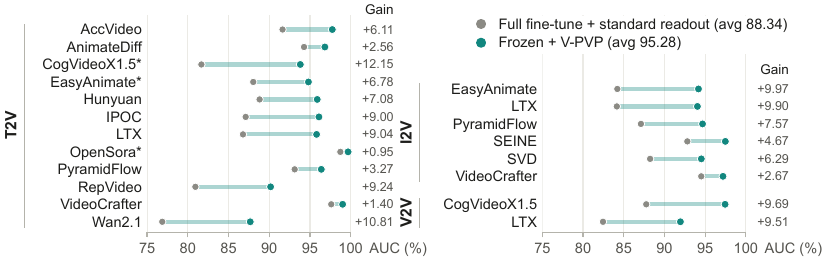}
    \caption{Per-generator AUC on \textbf{AIGVDBench} of fully fine-tuned VideoMAE with the standard readout (gray) and frozen VideoMAE with V-PVP (green), with the gain of V-PVP on the right. $^*$ marks the fake training generators.}
    \label{fig:generator_gains}
\end{figure}
A short temporal convolutional head reads $\{x_t\}_{t=1}^{T-1}$ into a single logit. Because $u_t$ and $w_t$ share the channel axis, every kernel position sees both signals at the same time step. A kernel size of three extends this to consecutive frame pairs, which matches the temporal scale of the inter frame inconsistencies V-PVP targets. A single anomalous transition is often not informative on its own, but a short window of them is. We then mean pool over time rather than learn a temporal attention, to avoid overfitting to the training generator's temporal position patterns.

The aggregator, the projections, and the head are trained from scratch. The backbone remains fully frozen. Full architecture details and training protocol are in Appendix~\ref{app:impl}.

\section{Experiments}
\label{sec:experiments}
\subsection{Setup}
\label{sec:exp:setup}

\paragraph{Datasets and protocol.}
We evaluate V-PVP on two cross-generator benchmarks. \textbf{AIGVDBench} \cite{ma2026your} contains a split spanning 20 generators across T2V, I2V, and V2V tasks. We adopt the three-generator training protocol of \cite{ma2026your}: OpenSora, CogVideoX1.5, and EasyAnimate serve as fake training sources. \textbf{GenVidBench-143k} \cite{ni2025genvidbench} adopts a cross-source protocol in which the training sources (Pika, VideoCrafterV2, ModelScope, T2V-Zero) and the test sources (MuseV, SVD, CogVideo, Mora) are fully disjoint. The primary metric is the mean per-generator AUC. Frames are sampled and preprocessed following the AIGVDBench protocol. The default backbone is VideoMAE~\cite{tong2022videomae} pretrained on Kinetics-400~\cite{kay2017kinetics} and kept fully frozen, with patch tokens extracted from the pre-final-norm position of the last transformer block. Full hyperparameters and dataset splits are in Appendix~\ref{app:pseudo}.

\paragraph{Baselines.}
We compare V-PVP against three families of baselines under the benchmark-specific cross-generator protocols. For AIGVDBench, all methods follow the protocol described above. For GenVidBench-143k, all methods follow its cross-source protocol with disjoint training and test generators. \textbf{AI-generated image detection models} \cite{tan2024rethinking,ojha2023towards,chen2025dual,yan2024orthogonal,liu2026mirror} classify each frame independently and aggregate by majority vote. \textbf{AI-generated video detection models} \cite{chen2024demamba,ma2025detecting,interno2025ai,zhang2025physics} are designed specifically for AIGV detection and use their original temporal modeling heads. \textbf{Video classification models} \cite{carreira2017quo,feichtenhofer2019slowfast,bertasius2021space,li2022uniformerv2,tong2022videomae} are adapted by full-parameter fine-tuning on the same training set, so that the comparison reflects each backbone's best achievable performance. All baselines are retrained on the corresponding training split following the unified evaluation protocol of each benchmark. Each method retains its original architecture and the hyperparameters recommended by the original authors, and only the training data is replaced. All methods are trained to convergence: training stops when the validation AUC improves by at most 0.5 points for three consecutive epochs, and the best-validation checkpoint is evaluated; ablations (Section~\ref{sec:exp:ablation}) use a fixed 5-epoch budget.

\subsection{Cross-Generator Comparison}
\label{sec:exp:main}
\begin{wraptable}{r}{0.5\textwidth}
    \vspace{-12pt}
    \centering
    \scriptsize
    \setlength{\tabcolsep}{2.5pt}
    \caption{Per-generator AUC on \textbf{GenVidBench-143k}. Best average in \textbf{bold}, second \underline{underlined}.}
    \label{tab:genvid}
    \begin{tabular}{lcccc|c}
        \toprule
        \textbf{Method} & MuseV & SVD & CogVideo & Mora & \textbf{AVG} \\
        \midrule
        \multicolumn{6}{c}{\textbf{AI-Generated Image Detection Models}} \\
        \midrule
        NPR \cite{tan2024rethinking} & 48.87 & 30.03 & 91.27 & 30.89 & 50.26 \\
        UnivFD \cite{ojha2023towards} & 65.79 & 55.37 & 97.40 & 96.46 & 78.76 \\
        Effort \cite{yan2024orthogonal} & 65.85 & 55.51 & 97.88 & 97.08 & 79.08 \\
        DDA \cite{chen2025dual} & 91.72 & 87.40 & 96.92 & 95.12 & 92.79 \\
        MIRROR \cite{liu2026mirror} & 77.29 & 74.59 & 59.93 & 91.42 & 75.81 \\
        \midrule
        \multicolumn{6}{c}{\textbf{AI-Generated Video Detection Models}} \\
        \midrule
        DeMamba \cite{chen2024demamba} & 88.46 & 73.05 & 98.80 & 99.80 & 90.03 \\
        DeCoF \cite{ma2025detecting} & 57.51 & 52.44 & 93.55 & 89.72 & 73.30 \\
        ReSTrav \cite{interno2025ai} & 86.04 & 82.12 & 96.97 & 95.99 & 90.28 \\
        NSG-VD \cite{zhang2025physics} & 78.26 & 63.99 & 72.83 & 61.80 & 69.22 \\
        \midrule
        \multicolumn{6}{c}{\textbf{Video Classification Models}} \\
        \midrule
        I3D \cite{carreira2017quo} & 91.91 & 87.17 & 94.85 & 97.12 & 92.76 \\
        SlowFast \cite{feichtenhofer2019slowfast} & 89.85 & 85.24 & 98.64 & 98.06 & \underline{92.95} \\
        TimeSformer \cite{bertasius2021space} & 89.06 & 85.92 & 91.72 & 98.00 & 91.18 \\
        UniFormerV2 \cite{li2022uniformerv2} & 66.66 & 64.42 & 59.63 & 88.78 & 69.87 \\
        VideoMAE \cite{tong2022videomae} & 82.14 & 81.78 & 94.55 & 95.89 & 88.59 \\
        \midrule
        \rowcolor[RGB]{210,225,250}
        \textbf{V-PVP (Ours)} & 90.27 & 88.63 & 97.79 & 98.30 & \textbf{93.75} \\
    \end{tabular}
\end{wraptable}
Tables~\ref{tab:aigvd_open} and~\ref{tab:genvid} report results on AIGVDBench and GenVidBench-143k. V-PVP reaches mga AUC of \textbf{95.28} on AIGVDBench and \textbf{93.75} on GenVidBench-143k, ranking first on both benchmarks. The backbone stays fully frozen and only about 0.5M parameters are trained.
Under the standard readout, some fully fine tuned video backbone in Table~\ref{tab:aigvd_open} sits below the best image level baseline, indicating that releasing backbone parameters alone does not unlock the temporal advantage promised by video pretraining. Our frozen backbone V-PVP exceeds end to end VideoMAE fine tuning (about 87M trainable parameters) with the standard readout by $+6.94$ AUC, ahead on all $20$ generators (Figure~\ref{fig:generator_gains}), and also surpasses the strongest image level baseline. With the backbone kept fully frozen, V-PVP simultaneously outperforms all video classification baselines, which are trained with full fine-tuning under their original readouts, as well as all image pretrained detectors, while training roughly two orders of magnitude fewer parameters. \textbf{This suggests that the standard aggregation readout is an important bottleneck for video backbones in AIGV detection.}

\subsection{Ablation Study}
\label{sec:exp:ablation}

We ablate V-PVP along two axes. The first looks inside V-PVP and asks how much each stream and the velocity gating contribute. The second looks outside and asks whether other readouts that drop the velocity prior can reach the same accuracy. All ablation runs use the frozen VideoMAE-K400 backbone with head-only training. All ablation rows including V-PVP use a fixed 5-epoch budget for fair comparison. Results are reported as mean AUC $\pm$ standard deviation over four random seeds. Except for this fixed epoch budget, the training hyperparameters are the same as the default V-PVP configuration in Appendix~\ref{app:pseudo}.

\paragraph{Per-stream contribution.}
Table~\ref{tab:streams} dissects V-PVP into its two streams. Going from $r$-only (MLPMean, 79.10) to $\bar z + r$ adds $+11.13$, most of which comes from the semantic information carried by the GAP pathway. Replacing $\bar z$ with the velocity-gated $\tilde z$ adds another $+4.11$ on top of an already channel-faithful stream. The middle two rows show that both components are needed: removing the $r$ stream drops AUC by $1.27$, and removing the velocity input from the scoring network drops it by $2.26$. Appendix~\ref{app:weights} visualizes the velocity-gated weights on success and failure cases.

\begin{table}[t!]
    \centering
    \small
    \caption{Ablation studies on V-PVP. Fixed 5-epoch ablation setting. Left: stream-level ablation. Right: alternative readouts on patch tokens or per-patch energy ($z$: patch tokens; $E$: $\|v_t^{(p)}\|_2$).}
    \label{tab:ablation}\label{tab:streams}\label{tab:spatial}
    \begin{subtable}[t]{0.61\textwidth}
        \centering
        \setlength{\tabcolsep}{4pt}
        
        \resizebox{\linewidth}{!}{        \begin{tabular}{lcc}
            \toprule
            \textbf{Configuration} & \textbf{Diff vs.\ V-PVP} & \textbf{AUC} \\
            \midrule
            $r$ alone, MLPMean                       & no $\tilde z$, GAP only      & 79.10\,$\pm$\,0.16 \\
            $\bar z + r$ (PVP)                       & $\tilde z \to$ GAP           & 90.23\,$\pm$\,0.21 \\
            $\tilde z$ alone (no $r$)                & no $r$ stream                & 93.07\,$\pm$\,0.35 \\
            $\tilde z + r$, no $\|v\|_2$ in scorer   & no velocity gating           & 92.08\,$\pm$\,1.84 \\
            \rowcolor[RGB]{210,225,250}
            \textbf{V-PVP ($\tilde z + r$)}          & ---                          & \textbf{94.34\,$\pm$\,0.32} \\
            \bottomrule
        \end{tabular}        }
    \end{subtable}
    \hfill
    \begin{subtable}[t]{0.38\textwidth}
        \centering
        \renewcommand{\arraystretch}{0.86}
        \setlength{\tabcolsep}{3pt}
        \resizebox{\linewidth}{!}{        \begin{tabular}{llr}
            \toprule
            \textbf{Head} & \textbf{Input} & \textbf{AUC} \\
            \midrule
            Mean (sanity)  & $E$     & 54.34\,$\pm$\,4.39 \\
            PatchMLP       & $z$     & 91.70\,$\pm$\,0.12 \\
            Conv2D         & $E$     & 68.61\,$\pm$\,0.70 \\
            Conv3D         & $E$     & 64.66\,$\pm$\,0.92 \\
            SpatialAttn    & $E$     & 67.37\,$\pm$\,0.98 \\
            \midrule
            \rowcolor[RGB]{210,225,250}
            \textbf{V-PVP (Ours)} & $z + \|v\|_2$ & \textbf{94.34\,$\pm$\,0.32} \\
            \bottomrule
        \end{tabular}        }
    \end{subtable}
\end{table}
\begin{figure}[!b]
    \centering
    \includegraphics[width=\linewidth]{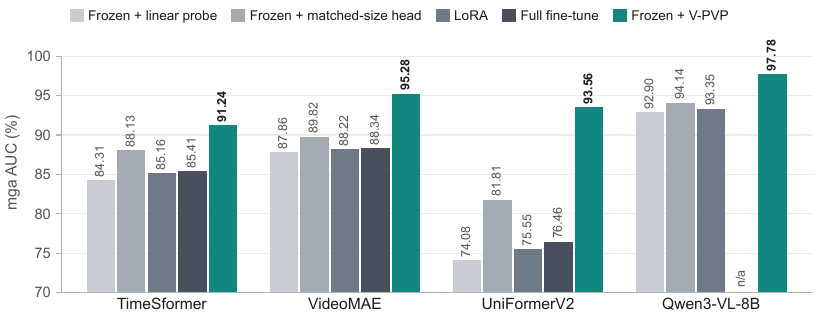}
    \caption{AUC comparison across different backbones and training strategies on \textbf{AIGVDBench}}
    \label{fig:backbone_auc}
\end{figure}
\paragraph{Alternative patch-level readouts.}
A natural question is whether a free-form head that keeps the patch dimension but drops the velocity prior could match V-PVP. We test this by training several such heads under identical supervision. Convolutional heads (Conv2D, Conv3D) and a spatial-attention head are applied to the per-patch energy map $E_t^{(p)} = \|v_t^{(p)}\|_2$, and a PatchMLP consumes the raw patch tokens directly as the most expressive alternative. Heads with more capacity to use the spatial dimension fit the training set more aggressively but generalise worse: PatchMLP and Conv2D reach near-perfect train AUC, yet their test mga AUC stays well below V-PVP. Without an explicit prior, the spatial pattern that supervision finds is generator-specific and does not transfer. V-PVP avoids this by anchoring patch selection to the generator-invariant quantity $\|v_t^{(p)}\|_2$, outperforming Conv2D by $25.73$ points ($94.34$ vs.\ $68.61$).

\subsection{Generality Across Backbones and Adaptation Schemes}
\label{sec:exp:backbone}

To test whether the bottleneck is the readout rather than the representation, we port V-PVP to several frozen video backbones and compare against the standard readout and adaptation alternatives.

We cover four ViT-based encoders with diverse pretraining objectives: TimeSformer~\cite{bertasius2021space}, VideoMAE~\cite{tong2022videomae}, UniFormerV2~\cite{li2022uniformerv2}, and the multimodal large model Qwen3-VL-8B~\cite{bai2025qwen3vltechnicalreport}. On every backbone we compare two groups of methods. The first keeps the backbone frozen and only trains a head, including a linear probe, a matched-size head, and our V-PVP. The second updates backbone parameters, including LoRA fine-tuning with rank 8 and full fine-tuning. Full fine-tuning is omitted on Qwen3-VL-8B due to its $\sim$8B parameter scale. All compared methods are trained to convergence with the stopping rule of Section~\ref{sec:exp:setup} under their respective default configurations; V-PVP uses the configuration in Appendix~\ref{app:impl}.

Figure~\ref{fig:backbone_auc} reports mga AUC on the AIGVDBench open-source split. V-PVP achieves the highest AUC on every backbone, surpassing all readout and adaptation alternatives while training roughly two orders of magnitude fewer parameters than full fine-tuning. The advantage holds across all four pretraining objectives, which supports the readout-bottleneck hypothesis: the bulk of the fine-tuning gain comes from adjusting how patch tokens are aggregated, not from modifying the representation itself. We also note that backbone tuning does not always help. On UniFormerV2, both LoRA and full fine-tuning stay below $80$, lower than the matched-size head with the backbone frozen. This suggests that updating backbone parameters can amplify overfitting to the training generator's spatial fingerprint rather than help. V-PVP avoids this failure mode by anchoring the readout to a generator-invariant quantity and reaches $93.56$ on the same backbone.

A natural follow-up question is whether the gains of V-PVP depend on a frozen backbone. We test this on VideoMAE by pairing V-PVP with LoRA at rank $8$ and with full fine-tuning, and compare against the same adaptation schemes under the standard readout. Table~\ref{tab:adaptation} reports mga AUC on AIGVDBench. The standard-readout column is nearly flat across adaptation regimes, indicating that backbone capacity has little outlet once patch tokens are collapsed by joint spatiotemporal pooling. V-PVP responds in the opposite way and improves over the standard readout under every adaptation scheme, with the largest margin reaching $8.9$ points under full fine-tuning. Even the frozen configuration with V-PVP, at roughly $0.5$M trainable parameters, surpasses full fine-tuning under the standard readout at roughly $87$M parameters. The right axis along which to invest capacity is therefore the readout rather than the backbone.

\begin{table}[t]
\centering
\caption{V-PVP composes with backbone adaptation. mga AUC on AIGVDBench with VideoMAE under three adaptation schemes. V-PVP improves over the standard readout under every scheme; the frozen V-PVP configuration already exceeds full fine-tuning under the standard readout at a fraction of the parameter cost.}
\label{tab:adaptation}
\begin{tabular}{lcccc}
\toprule
Adaptation & Trainable params & Standard readout & V-PVP & $\Delta$ \\
\midrule
Frozen     & $\sim$0.5\,M  & 89.82 & \textbf{95.28} & $+5.5$ \\
LoRA ($r{=}8$) & $\sim$1\,M    & 88.22 & \textbf{95.06} & $+6.8$ \\
Full fine-tune & $\sim$87\,M   & 88.34 & \textbf{97.21} & $+8.9$ \\
\bottomrule
\end{tabular}
\end{table}

\section{Conclusion}
\label{sec:conclusion}
We identify the readout layer as the principal bottleneck that prevents video pretrained backbones from realizing their temporal advantage on AIGV detection. The standard CLS or GAP aggregator performs a joint spatiotemporal and semantic collapse over patch tokens, which erases inter patch relations and dilutes per channel velocity magnitudes, the cues that AIGV detection most reliably depends on.

Building on this diagnosis, we propose V-PVP, a lightweight readout that keeps the backbone fully frozen and replaces only the aggregation layer with two parallel streams over the patch velocity field: a velocity gated aggregation $\tilde z_t$ that carries inter patch heterogeneity into the pooled output, and a channel faithful stream $r_t$ that preserves per channel magnitudes. With only about 0.5M trainable parameters, V-PVP consistently surpasses same backbone readouts across multiple ViT based backbones and two cross generator benchmarks, and with frozen VideoMAE it reaches 95.28 mga AUC on AIGVDBench, above full fine tuning under the standard readout. For AIGV detection, the right axis to invest capacity is therefore the readout rather than the backbone.

\paragraph{Limitations.}
V-PVP has several known limitations. First, generators that deliberately mimic real video statistics, for example by injecting inter frame jitter or preserving high frequency spatial detail, can shrink the inter patch separability of $\lVert v_t^{(p)} \rVert_2$ and drive velocity gating toward uniform attention. Second, very short clips with $T<4$ leave too few temporal samples of patch velocity and increase the variance of $\tilde z_t$ and $r_t$. Third, when the backbone has not been pretrained on a distribution close to the target domain, the frozen patch tokens may be too weak to support readout only learning, and LoRA or full fine tuning would be more appropriate. Finally, the readout bottleneck still exists on CNN based backbones but in a softened form, since internal convolutions already perform part of the spatial aggregation before the readout, partially absorbing the patch level evidence that V-PVP relies on. Appendix~\ref{app:cnn} confirms this on I3D-R50 and SlowFast-R50, but a thorough study of CNN specific readout design is left to future work.

\begin{ack}
The work was supported by the National Key R\&D Program of China under Grant \mbox{2021ZD0201300}, the National Natural Science Foundation of China under Grant \mbox{62406118}, the Hubei Provincial Science and Technology Innovation Program (Key Research and Development Project, Grant No.~\mbox{2026BCB099}), the Postdoctoral Project of Hubei Province under Grant Number \mbox{2024HBBHCXA017}, the Foundation for Key Research and Development Program Project of Wuhan City under Grant \mbox{2025051202030404}, and the Hubei Provincial Technology Innovation Program under Grant \mbox{2026CSA035}.
\end{ack}

\bibliographystyle{unsrtnat}
\bibliography{main}

\newpage
\appendix

\section{Implementation and Comparison Protocol}
\label{app:impl}

This appendix gives the implementation details behind the controlled comparisons in Section~\ref{sec:exp:setup}. The goal of the protocol is to isolate where performance gains come from: the frozen video representation, the readout that aggregates patch tokens, or adaptation of the backbone itself.

\paragraph{Shared protocol.}
Unless otherwise specified, the default V-PVP setting uses VideoMAE-Base pretrained on Kinetics-400. The backbone receives $T=16$ frames at $224\times224$ resolution and exposes $P=196$ spatial patch tokens with hidden width $D=768$. We tap tokens from the output of the last transformer block, layer 11, at the pre-final-norm position. The backbone is frozen end to end: no gradients flow through it, no statistics are updated, and no AIGV labels are used to modify its representation. This makes the readout the only trainable component in the default setting.

\paragraph{Compute resources.}
All experiments were conducted on a single workstation with 4 NVIDIA A100 80\,GB PCIe GPUs using PyTorch and CUDA. Each training run uses one A100 GPU without multi-GPU or distributed training; multiple runs are executed in parallel across the available GPUs. Since the video backbones are frozen, training only updates the lightweight readout head. The wall-clock cost per backbone configuration is approximately 3--8 GPU-hours, including evaluation. Across the main backbone comparisons, method variants, seed sweeps, and ablations, the reported experiments required on the order of 400--600 A100-80GB GPU-hours in total.

For V-PVP in the main comparisons, we train only the readout with binary cross-entropy on the clip-level logit and use a positive-class weight $w_{\text{pos}}=N_{\text{neg}}/N_{\text{pos}}$ to account for the real-to-fake imbalance in the training split. Optimization uses AdamW with learning rate $10^{-3}$, weight decay $10^{-4}$, cosine annealing, batch size 192, and mixed precision. The readout parameters are initialized from scratch. For the main comparisons in Tables~\ref{tab:aigvd_open} and~\ref{tab:genvid}, V-PVP is trained with this configuration to convergence: training stops once the validation AUC fails to improve by more than 0.5 percentage points for three consecutive epochs, and the checkpoint with the highest validation AUC is selected for final evaluation. Baseline methods follow their default model-specific configurations and use the same stopping rule, as described in Section~\ref{sec:exp:setup}.

\paragraph{V-PVP instantiation.}
V-PVP replaces only the aggregation layer on top of the frozen patch tokens. For each adjacent frame pair, we compute the patch velocity $v_t^{(p)}=z_{t+1}^{(p)}-z_t^{(p)}$. The velocity-gated branch scores each patch with a two-layer MLP of shape $D+1\to128\to1$, whose input is the patch token $z_t^{(p)}$ concatenated with the scalar velocity magnitude $\lVert v_t^{(p)}\rVert_2$. A softmax over patches produces the aggregation weights used in Equation~\ref{eq:vgap}. In parallel, the channel-faithful velocity branch computes the mean absolute patch velocity in Equation~\ref{eq:rt}, preserving per-channel magnitude before any signed cancellation can occur.

The two streams are projected independently by linear maps $W_z,W_r:\mathbb{R}^{D}\to\mathbb{R}^{H/2}$ with $H=256$, and their outputs are concatenated into $x_t\in\mathbb{R}^{H}$ for each frame pair. The temporal head then consumes the length-$(T-1)$ sequence $\{x_t\}_{t=1}^{T-1}$ with
\[
\mathrm{GELU}\rightarrow \mathrm{Conv1d}(H,H,k=3)\rightarrow \mathrm{GELU}
\rightarrow \mathrm{MeanPool}\rightarrow \mathrm{Dropout}(0.3)\rightarrow \mathrm{Linear}(H\to1).
\]
The temporal convolution is deliberately lightweight: after velocity-gated aggregation and channel-faithful profiling have extracted the local temporal evidence, the head only needs to smooth adjacent frame-pair features before clip-level pooling. The full readout contains 492,674 trainable parameters, about $0.57\%$ of the VideoMAE-Base backbone.

\paragraph{Ablation protocol.}
For the ablation study in Table~\ref{tab:ablation}, all variants use the same V-PVP optimization settings above except that training is fixed to 5 epochs for controlled relative comparison. Under this fixed-budget setting, the full V-PVP obtains 94.34 AUC; the main comparison result in Table~\ref{tab:aigvd_open} is obtained with the convergence-based stopping rule above.

\paragraph{Readout-only baselines.}
The frozen-backbone baselines are organized by the question each one answers rather than by architecture name. The first group tests whether a stronger head on the standard global representation is sufficient. These baselines operate on the GAP-pooled or CLS-style sequence $c$ exposed by the action-recognition pipeline and include LinearMean, MLPMean, TConv1D, and TinyXfmr. Since they see the same representation as a conventional video classifier, their gap to V-PVP measures how much information is lost by the standard readout interface.

The second group tests how far the channel-faithful velocity stream can go by itself. We apply the same LinearMean, MLPMean, TConv1D, and TinyXfmr heads to $r$ alone, where $r$ is the per-channel mean-of-magnitudes velocity readout without the velocity-gated token branch. This isolates the discriminative value of velocity magnitude before fusion with token-level semantic evidence.

The third group gives non-V-PVP heads access to multiple pooled signals. These joint $rcs$ baselines concatenate the velocity readout $r$, the CLS-style global summary $c$, and a per-frame spatial-mean token $s$. They are the closest readout-only alternatives that still combine velocity evidence with a global token summary, but without input-adaptive velocity-gated patch selection.

The final group asks whether preserving the spatial dimension without the V-PVP prior is enough. We route the per-patch energy map $E_t^{(p)}=\lVert v_t^{(p)}\rVert_2$ through Conv2D, Conv3D, and SpatialAttn heads. These heads can in principle learn spatially structured evidence from data, but they do not explicitly condition patch aggregation on the velocity-gated mechanism in Equation~\ref{eq:vgap}. Their role is therefore to test whether unconstrained spatial supervision can replace the structured readout design.

\paragraph{Backbone-adaptation baselines.}
To separate readout design from representation adaptation, we also compare against methods that change how much of the backbone is trained. The linear probe and matched-size head keep the backbone frozen and vary only the capacity of the classifier. LoRA fine-tuning, with rank 8 in our experiments, updates a small set of backbone adaptation parameters while retaining the standard readout. Full fine-tuning updates the entire video backbone with the same task supervision and serves as the high-capacity adaptation reference. Full fine-tuning is omitted for Qwen3-VL-8B because of its parameter scale. For the backbone comparison in Figure~\ref{fig:backbone_auc}, all methods are trained to convergence with the same validation-based stopping rule under their respective configurations; V-PVP uses the same readout-only configuration described above. These baselines answer whether AIGVDBench supervision is better spent modifying the representation or replacing the aggregation layer.
\section{Pseudocode and Experimental Configuration}
\label{app:pseudo}

\subsection{Inference pseudo-code}

Algorithm~\ref{alg:vpvp} writes out the inference path. The forward pass is identical to the training-time one; the only differences are that gradients are not tracked and the clip-level augmentation is replaced with center cropping.

\begin{algorithm}[H]
\caption{V-PVP Inference}\label{alg:vpvp}
\begin{algorithmic}[1]
\Require Video $x$, frozen backbone $F$, trainable parameters $\theta_{\mathrm{MLP}}$, $W_z$, $W_r$, $\theta_{\mathrm{head}}$
\Ensure Probability $P(\text{fake})$

\State Sample $T$ frames from $x$ and apply preprocessing
\State $Z \leftarrow F(x)$ \Comment{$Z \in \mathbb{R}^{P \times T \times D}$, no gradient through $F$}

\For{$t = 1, \dots, T{-}1$}
    \For{$p = 1, \dots, P$}
        \State $v_t^{(p)} \leftarrow z_{t+1}^{(p)} - z_t^{(p)}$ \Comment{Patch velocity}
        \State $s_t^{(p)} \leftarrow \mathrm{MLP}_\theta\!\bigl([z_t^{(p)},\; \lVert v_t^{(p)} \rVert_2]\bigr)$ \Comment{Attention score}
    \EndFor
    \State $\alpha_t^{(p)} \leftarrow \mathrm{softmax}_p\!\bigl(s_t^{(p)}\bigr)$ \Comment{Patch weights}
    \State $\tilde{z}_t \leftarrow \sum_{p} \alpha_t^{(p)}\, z_t^{(p)}$ \Comment{Velocity-gated aggregation}
    \State $r_t \leftarrow \frac{1}{P} \sum_{p} \bigl| v_t^{(p)} \bigr|$ \Comment{Channel-faithful stream}
    \State $x_t \leftarrow \bigl[W_z\, \tilde{z}_t \;;\; W_r\, r_t\bigr]$ \Comment{Project and concatenate}
\EndFor

\State $h \leftarrow \mathrm{MeanPool}_t\!\bigl(\mathrm{GELU}\bigl(\mathrm{Conv1D}_{k=3}\bigl(\mathrm{GELU}(\{x_t\})\bigr)\bigr)\bigr)$
\State $\mathrm{logit} \leftarrow \mathrm{Linear}\!\bigl(\mathrm{Dropout}(h)\bigr)$
\State \Return $\sigma(\mathrm{logit})$

\end{algorithmic}
\end{algorithm}

\subsection{Experimental configuration}

Table~\ref{tab:expcfg} consolidates the default configuration used by V-PVP.
For the main comparisons in Tables~\ref{tab:aigvd_open} and~\ref{tab:genvid}, V-PVP is trained
to convergence: training stops once the validation AUC fails to improve by more
than 0.5 percentage points for three consecutive epochs, and the checkpoint with
the highest validation AUC is selected for final evaluation. Other compared
methods keep their default model-specific configurations and use the same
stopping rule. For
the ablation study in Table~\ref{tab:ablation}, all ablation variants use a fixed
5-epoch budget for controlled relative comparison; all other hyperparameters are
the same as the default V-PVP configuration below.

\begin{table}[H]
\centering
\caption{Default V-PVP training configuration. Values are shared across the
reported V-PVP runs unless explicitly stated otherwise. The number of training
epochs is not fixed for the main comparisons; training stops once the validation
AUC fails to improve by more than 0.5 percentage points for three consecutive
epochs.}
\label{tab:expcfg}
\small
\begin{tabular}{ll}
\toprule
\textbf{Setting} & \textbf{Value} \\
\midrule
\multicolumn{2}{l}{\textit{Backbone}} \\
Architecture & VideoMAE-Base \\
Pretraining & Kinetics-400 \\
Frozen & Yes  \\
Token tap & Last transformer block (layer 11), pre-final-norm \\
Spatial patch tokens ($P$) & 196 \\
Hidden width ($D$) & 768 \\
\midrule
\multicolumn{2}{l}{\textit{Input}} \\
Frames per clip ($T$) & 16 \\
Spatial resolution & $224 \times 224$ \\
\midrule
\multicolumn{2}{l}{\textit{Trainable readout}} \\
Patch scorer MLP & $D{+}1 \to 128 \to 1$ \\
Channel projection width ($H$) & 256 \\
Sequence head & GELU $\to$ Conv1D($k{=}3$) $\to$ GELU $\to$ MeanPool $\to$ Dropout(0.3) $\to$ Linear \\
Total trainable parameters & 492,674 ($\sim$0.57\% of backbone) \\
\midrule
\multicolumn{2}{l}{\textit{Optimization}} \\
Loss & Binary cross-entropy with class weighting \\
Optimizer & AdamW \\
Learning rate & $10^{-3}$ \\
Weight decay & $10^{-4}$ \\
LR schedule & Cosine annealing \\
Batch size & 192 \\
Mixed precision & Enabled \\
Initialization & Readout from scratch; backbone frozen \\
Stopping criterion & Validation AUC gain $\le$ 0.5 pt for 3 consecutive epochs \\
Checkpoint selection & Highest validation AUC \\
\midrule
\multicolumn{2}{l}{\textit{Data (AIGVDBench)}} \\
Train fake sources & Open-Sora, CogVideoX1.5, EasyAnimate \\
Evaluation split & 20 generators, 3k fake videos per generator, plus 3k real videos \\
Reported metric & Mean per-generator AUC \\
\bottomrule
\end{tabular}
\end{table}
\section{V-PVP on CNN Backbones}
\label{app:cnn}

The readout bottleneck identified in Section~\ref{sec:motivation} is most directly motivated on ViT class backbones, where every patch token is anchored to a fixed input region and the readout is the single place where the spatial axis collapses. CNN class backbones offer a useful additional test case, since their layer wise convolutions already perform part of the spatial aggregation before the readout. If our diagnosis is correct, two predictions follow. V-PVP should still bring a clear improvement on CNN backbones, because the final readout still collapses what remains of the spatial axis. And the improvement should be smaller than on ViT backbones, because part of the aggregation V-PVP targets has already happened implicitly. We verify both on I3D-R50 and SlowFast-R50.

The extension to CNN feature maps is straightforward. Given a feature map of shape $(B, C, T, H, W)$, we reshape it to $(B, P, T, D)$ with $P = H \times W$ and $D = C$, treating each spatial cell as a pseudo patch token. The patch velocity in Equation~\ref{eq:velocity} and the rest of the V-PVP head apply without modification. We tap from the last residual stage, where $H = W = 7$ and $D = 2048$ give $P = 49$ patch tokens; for SlowFast we use the slow pathway with $T = 8$. The backbones are pretrained on Kinetics-400 and kept fully frozen, with protocol and splits identical to Section~\ref{sec:experiments}.

\begin{table}[h]
\centering
\caption{V-PVP versus the standard readout on frozen CNN backbones, mga AUC on AIGVDBench.}
\label{tab:cnn_results}
\begin{tabular}{lccccc}
\toprule
Backbone & Adaptation & Trainable params & Std readout & V-PVP & $\Delta$ \\
\midrule
I3D-R50 & Frozen & $\sim$1.0\,M & 83.62 & \textbf{86.53} & $+2.91$ \\
SlowFast-R50 & Frozen & $\sim$1.0\,M & 82.35 & \textbf{85.06} & $+2.71$ \\
\bottomrule
\end{tabular}
\end{table}

As shown in Table~\ref{tab:cnn_results}, V-PVP delivers consistent gains on CNN backbones, improving mga AUC by $+2.91$ on I3D-R50 and $+2.71$ on SlowFast-R50 with about 1M trainable parameters (the wider $D = 2048$ input enlarges the patch scorer and the two projections), confirming that the readout aggregation is also a bottleneck on this family. The gains are smaller than on every ViT backbone in Figure~\ref{fig:backbone_auc}, where frozen V-PVP improves over the frozen linear probe by $+4.88$ to $+19.48$ AUC and over the frozen matched-size head by $+3.11$ to $+11.75$ AUC, matching the second prediction.

This cross family pattern provides additional evidence for our diagnosis. If the bottleneck were located inside the backbone representation rather than at the readout, we would expect comparable gains across architectures, since V-PVP modifies neither. Instead, ViT backbones that carry patch level evidence intact up to the readout gain the most, while CNN backbones whose convolutional stages have already smoothed part of the spatial structure gain less. The size of the improvement tracks how much patch level evidence survives until the readout, exactly as the readout bottleneck hypothesis predicts.

We further verify this on I3D-R50 by recomputing the diagnostics of Section~\ref{sec:motivation}. The Loss~1 spectral condition number yields Cohen's $d = 0.42$, below the $0.64$ on VideoMAE: the inter patch signal does not vanish but is partially absorbed by the convolutional stages. Loss~2 shows a stronger form of the same effect. On VideoMAE, pooling lowers the separability of the velocity energy from $d = 0.69$ pre pool to $0.41$ post pool, which is the dilution that $r_t$ undoes. On I3D the order is reversed: the pre pool energy is essentially non-discriminative ($d \approx 0$) while the post pool energy reaches $d = 0.36$, so pooling dilutes no separability and there is little for $r_t$ to recover. Both diagnostics leave less for V-PVP to recover on I3D, matching the smaller AUC improvement in Table~\ref{tab:cnn_results}.
\section{Derivation and Validation}

This appendix grounds both V-PVP streams in the structure of the two losses identified in the main text, presenting each stream as the natural endpoint of a chain of design decisions rather than an isolated engineering choice. The three steps below progressively constrain the form of the repair operator, and the empirical analyses in Section~\ref{app:validation} verify the key claims along this chain. To make the diagnosis applicable beyond a single readout, we first group the common readouts into a single family, lifting the diagnosis from a GAP-specific observation to a family-level statement.

\subsection{Derivation from the Two Losses}
\label{app:derivation}

\subsubsection{Formal Setup and the Standard Readout Family}

The frozen backbone produces patch tokens $\{z_t^{(p)}\}$ with $z_t^{(p)} \in \mathbb{R}^D$, and the patch velocity is
\begin{equation}
v_t^{(p)} = z_{t+1}^{(p)} - z_t^{(p)}.
\end{equation}

\paragraph{Definition 1 (Standard Readout Family $\mathcal{S}$).}
A readout $f$ belongs to the family $\mathcal{S}$ if for any input $\{z^{(p)}\}_{p=1}^{P}$ there exist input-dependent weights $\omega^{(p)} \ge 0$ with $\sum_p \omega^{(p)} = 1$ such that
\begin{equation}
f(\{z^{(p)}\}) = \sum_p \omega^{(p)} z^{(p)}.
\end{equation}
The defining property of $\mathcal{S}$ is that the patch axis is collapsed in a single step into a $D$-dimensional output. GAP corresponds to the constant weights $\omega^{(p)} = 1/P$. The CLS token uses input-dependent attention weights but still compresses the $P \times D$ patch tensor into a single $D$-dimensional vector at the readout interface. Fixed-query attention pooling shares the same structure. We use $\mathcal{S}$ as an abstraction of these single-vector patch-collapsing readouts, with the common feature that the patch axis is folded into $D$ dimensions at once.

\subsubsection{Loss 1: Loss and Recovery of Inter-Patch Relations}

The first question is whether some choice of $\omega^{(p)}$ within $\mathcal{S}$ can preserve inter-patch relations.

\paragraph{Step 1: Inter-patch relations cannot be preserved by any $f \in \mathcal{S}$.}
Arrange patch velocities as $V_t \in \mathbb{R}^{P \times D}$ with each row being one patch velocity vector. The inter-patch relations are encoded in the patch-dimension Gram matrix
\begin{equation}
G_t = V_t V_t^{\top} \in \mathbb{R}^{P \times P}.
\end{equation}
Since $G_t$ lives in a $P^2$-dimensional space and any $f \in \mathcal{S}$ folds the patch axis into a $D$-dimensional vector, $f$ cannot in general be an injective representation of $G_t$. Selected projections of $G_t$ may remain, but the full inter-patch relation matrix is not recoverable after the collapse.

This conclusion is a structural property of the family $\mathcal{S}$ and holds equally for the uniform weights of GAP, the input-dependent attention weights of CLS, and the position-heterogeneous weights of fixed-query pooling.

The second question is the direction of repair. Lifting the output dimension to $D^2$ via second-order covariance pooling preserves $G_t$ but increases parameter and downstream fusion cost substantially. An alternative is to accept the $D$-dimensional capacity limit and look for a discriminative projection of the inter-patch relations within this $D$-dimensional space. AIGV detection eventually requires only a scalar discriminant function, so a discriminative $D$-dimensional projection is sufficient. We take this second path.

The remaining question is whether merely changing how $\omega^{(p)}$ is generated can carry such a projection within the $D$-dimensional output. We compare an input-adaptive weighting against an arbitrary baseline in $\mathcal{S}$ to see the geometric meaning of the difference.

\paragraph{Step 2: Input-adaptive weights introduce a content-dependent residual on top of any baseline in $\mathcal{S}$.}
For any baseline $f_0 \in \mathcal{S}$ with weights $\omega^{(p)}$, let $z_\omega := f_0(\{z^{(p)}\})$. For any input-adaptive weights $\alpha^{(p)}$ with $\sum_p \alpha^{(p)} = 1$, a direct expansion using $\sum_p \omega^{(p)} (z^{(p)} - z_\omega) = 0$ yields
\begin{equation}
\sum_p \alpha^{(p)} z^{(p)} = z_\omega + C(\alpha, \omega, z), \qquad C(\alpha, \omega, z) := \sum_p (\alpha^{(p)} - \omega^{(p)}) (z^{(p)} - z_\omega).
\label{eq:residual}
\end{equation}

The decomposition shows that input-adaptive weighting equals the baseline output plus a content-dependent residual vector $C \in \mathbb{R}^D$. The term $C$ does not attempt to recover the full $P \times P$ relation matrix, which exceeds the capacity of the $D$-dimensional output. Instead it encodes a content-dependent projection of the relation structure along the weight deviation direction $\alpha^{(p)} - \omega^{(p)}$. The decomposition holds for any baseline in $\mathcal{S}$, and the choice of $f_0$ only shifts the reference point against which $C$ is measured.

The remaining question is what $\alpha$ should be conditioned on. From the perspective of signal separability, the inter-patch spectral condition number $\kappa_{\text{spec}}$ of $\|v^{(p)}\|$ achieves Cohen's $d \approx 0.64$ between real and fake videos and is therefore an empirically validated separable quantity. Conditioning the scorer explicitly on $\|v^{(p)}\|$ allows the weight deviation $\alpha^{(p)} - \omega^{(p)}$ to anchor directly to this temporal cue. From the perspective of generator invariance, a widely observed failure mode in AIGV detection is that detectors memorize the low-level spatial statistics of the training generator and fail on unseen ones \cite{tan2024rethinking, ojha2023towards, chen2025dual}. Since $z^{(p)}$ carries such static appearance statistics, a scorer relying only on $z^{(p)}$ tends to align $\alpha$ with the semantic fingerprint of the training generator. The patch velocity $\|v^{(p)}\|$ is obtained from a first-order temporal difference and naturally filters out the static spatial component, providing a physical anchor for the notion of anomaly that depends weakly on single-frame appearance. The two perspectives converge on the same answer that the conditioning input should include $\|v^{(p)}\|$.

Concatenating $z^{(p)}$ and $\|v^{(p)}\|$ gives the simplest conditioning input. Softmax is the simplest normalization satisfying $\sum_p \alpha^{(p)} = 1$, and the simplest scoring function is a small MLP. We therefore obtain
\begin{equation}
\alpha_t^{(p)} = \mathrm{softmax}_p \! \left( \mathrm{MLP} \! \left( [z_t^{(p)} ; \|v_t^{(p)}\|_2] \right) \right), \qquad \tilde{z}_t = \sum_p \alpha_t^{(p)} z_t^{(p)}.
\end{equation}
The first stream is therefore the minimal repair operator that arises from combining Step 1, the residual decomposition of Step 2, and the two constraints on the conditioning variable.

\subsubsection{Loss 2: Dilution and Recovery of Per-Channel Magnitudes}

We follow the same logic as in the previous subsection and first establish a family-level bound.

\paragraph{Step 3: Per-channel magnitudes are bounded above by the pre-summation magnitude on $\mathcal{S}$.}
For any $f \in \mathcal{S}$ and any channel $d$, applying Jensen's inequality to the convex function $|\cdot|$ with weighted probability measure $\omega^{(p)}$ gives
\begin{equation}
\sum_p \omega^{(p)} \, |v^{(p)}[d]| \ge \left| \sum_p \omega^{(p)} \, v^{(p)}[d] \right|,
\label{eq:jensen}
\end{equation}
with equality if and only if $\{v^{(p)}[d]\}$ is sign-consistent or zero on the support $\{p : \omega^{(p)} > 0\}$. The argument is performed sample-wise so that input-dependent $\omega^{(p)}$ is allowed.

Define the formal quantity for Loss 2 as $\mathcal{L}_2^{\mathcal{S}}(d) := \sum_p \omega^{(p)} |v^{(p)}[d]| - |\sum_p \omega^{(p)} v^{(p)}[d]| \ge 0$, which is a family-level lower bound independent of whether $f$ is GAP, CLS, or fixed-query pooling. The key implication is that any operator that first performs a patch-weighted sum and then applies a channel-wise non-linearity has already lost $\mathcal{L}_2^{\mathcal{S}}$ to Jensen's inequality. Recovery requires placing the element-wise non-linearity before the patch summation, which is the only path out of $\mathcal{S}$.

We therefore consider the candidate family
\begin{equation}
g_\phi(v)[d] = \frac{1}{P} \sum_p \phi(v^{(p)}[d]),
\end{equation}
where $\phi : \mathbb{R} \to \mathbb{R}_{\ge 0}$ is an element-wise function. Three design criteria narrow the choice of $\phi$ to a small set of candidates. The first is suppression of cancellation, requiring $\phi(x) > 0$ for $x \neq 0$, which directly recovers $\mathcal{L}_2^{\mathcal{S}}$ and rules out the identity map and ReLU (the latter discards negative magnitudes). The second is sign symmetry, $\phi(x) = \phi(-x)$, since generator artifacts have no systematic sign preference along the channel direction. The third is statistical robustness. Treating $g_\phi$ as a sample-mean estimator of $\phi(V[d])$, where $V[d]$ denotes the distribution of patch velocities on channel $d$, the variance is $\mathrm{Var}[\phi(V[d])]/P$ but the constant $\mathrm{Var}[\phi(V[d])]$ depends on different moments of the noise distribution for different $\phi$. If $V[d] \sim \mathcal{N}(0, \sigma^2)$, then $\mathrm{Var}[|V[d]|] = (1 - 2/\pi) \sigma^2$ depends only on the second moment, while $\mathrm{Var}[V[d]^2] = 2 \sigma^4$ depends on the fourth moment and is sensitive to changes in noise scale. Since motion noise variance $\sigma^2$ fluctuates across videos, a $\phi$ whose variance depends only on the second moment yields a more stable feature. This property is formalized as first-order homogeneity $\phi(\lambda x) = |\lambda| \phi(x)$.

The simplest element-wise scalar form satisfying all three criteria is $\phi(x) = c |x|$ with $c > 0$, obtained by taking $\lambda > 0$ and $x > 0$ in the third criterion to give $\phi(x) = \phi(1) \cdot x$, then applying the second to give $\phi(-x) = \phi(1) \cdot |-x|$, and the first to ensure $\phi(1) > 0$. Smooth surrogates such as $\sqrt{x^2 + \varepsilon}$ satisfy the same criteria but introduce an additional hyperparameter, so we adopt $|\cdot|$ for its parameter-free simplicity. The minimal repair operator for Loss 2 is therefore
\begin{equation}
r_t = \frac{1}{P} \sum_p |v_t^{(p)}|.
\end{equation}
Section~\ref{app:phi_compare} verifies this choice empirically against three boundary alternatives that each violate one criterion.

\subsection{Empirical Analyses}
\label{app:validation}

The derivation makes two specific claims at key nodes. The three criteria narrow $\phi$ to $|\cdot|$ as a parameter-free choice for Loss 2, and the residual decomposition in Step 2 should manifest as differentiated contributions of the two streams across sample subsets. Both claims are indirectly supported by the overall accuracy reported in the main text, but accuracy numbers alone cannot distinguish the derivation path from alternative explanations consistent with the same numbers. This section provides two analyses that tighten the intermediate steps. All experiments use the AIGVDBench Open-Source split with 3k real and 60k fake test videos and the frozen VideoMAE-K400 backbone. The discriminability metric is Cohen's $d$ between real and fake.

\subsubsection{Boundary Comparison of the Three Criteria}
\label{app:phi_compare}

Reproducing the gain of $r_t$ over $\bar{v}_t$ alone only verifies that placing a non-linearity before patch summation is effective, not that $|\cdot|$ is preferable to similar candidates. A more discriminative test comes from the boundaries of the three criteria. If any criterion is relaxed, the form of $\phi$ deviates from $|\cdot|$ and the corresponding $r_t$ variant should exhibit a measurable drop in discriminability.

We compare four choices of $\phi$. The choice $|\cdot|$ satisfies all three criteria. ReLU violates the first by mapping negative magnitudes to zero. $x^2$ violates the third since its variance depends on the fourth moment. The identity map uses no pre-summation non-linearity and corresponds to the velocity-domain baseline $\bar{v}_t \in \mathcal{S}$. For each choice we compute $r_\phi = \frac{1}{P} \sum_p \phi(v^{(p)})$, average across time, and report the mean and standard deviation of per-channel Cohen's $d$ across all $D = 768$ channels.

\begin{table}[h]
\centering
\caption{Per-channel discriminability of four $\phi$ candidates. Relative gap is computed against $\phi = |\cdot|$.}
\label{tab:phi_compare}
\begin{tabular}{lccc}
\toprule
$\phi$ & Violation & Per-channel $d$ (mean $\pm$ std) & Relative gap \\
\midrule
$|\cdot|$ (V-PVP) & none & $0.372 \pm 0.092$ & --- \\
$\mathrm{ReLU}$ & first criterion & $0.340 \pm 0.102$ & $-9\%$ \\
$x^2$ & third criterion & $0.321 \pm 0.101$ & $-14\%$ \\
$\mathrm{id}$ ($\bar{v}_t$) & no pre-summation non-linearity & $0.153 \pm 0.111$ & $-59\%$ \\
\bottomrule
\end{tabular}
\end{table}

The ordering $|\cdot| > \mathrm{ReLU} > x^2 \gg \mathrm{id}$ matches the prediction. The identity map collapses by 59 percent, a direct empirical instantiation of the Jensen lower bound on $\mathcal{S}$ since opposite-signed values cancel along each channel before the absolute value is taken. ReLU loses the negative magnitude component and falls by 9 percent, while $x^2$ inflates the variance through fourth-moment dependence and falls by 14 percent. The advantage of $|\cdot|$ does not come from the general structure of placing a non-linearity before summation, but from the specific form selected by the three criteria.

\subsubsection{Differentiated Contributions of the Two Streams}
\label{app:bucket}

The two streams are obtained from the two losses through independent derivation paths and share no intermediate reasoning. The residual term $C$ in Step 2 depends explicitly on the semantic content of $z^{(p)} - z_\omega$, while $r_t$ depends only on $|v^{(p)}|$ and discards all semantic content. The formal independence of the two streams is therefore clear. The ablation in the main text shows that $\tilde{z}$ alone reaches 93.07 AUC and $\tilde{z} + r$ reaches 94.34, a gap of 1.27. This indicates that $\tilde{z}$ already absorbs a substantial portion of the discriminative signal carried by $r$ in terms of overall accuracy, so the differentiated contributions are best observed on finer sample subsets and on sample-level error patterns rather than on aggregate accuracy.

We examine this from two angles. The first is the sample-level correlation of the errors of the two streams. We train head-only single-stream models $M_z$ using $\tilde{z}_t$ and $M_r$ using $r_t$ with three random seeds each, holding all other configurations consistent with V-PVP. On the test set we record per-sample binary error labels $e_z, e_r \in \{0, 1\}$ thresholded at sigmoid output 0.5, and report the Pearson $\phi$ coefficient and the conditional error rate $P(e_r = 1 \mid e_z = 1)$ averaged across seeds.

\begin{figure}[p]
    \centering
    \includegraphics{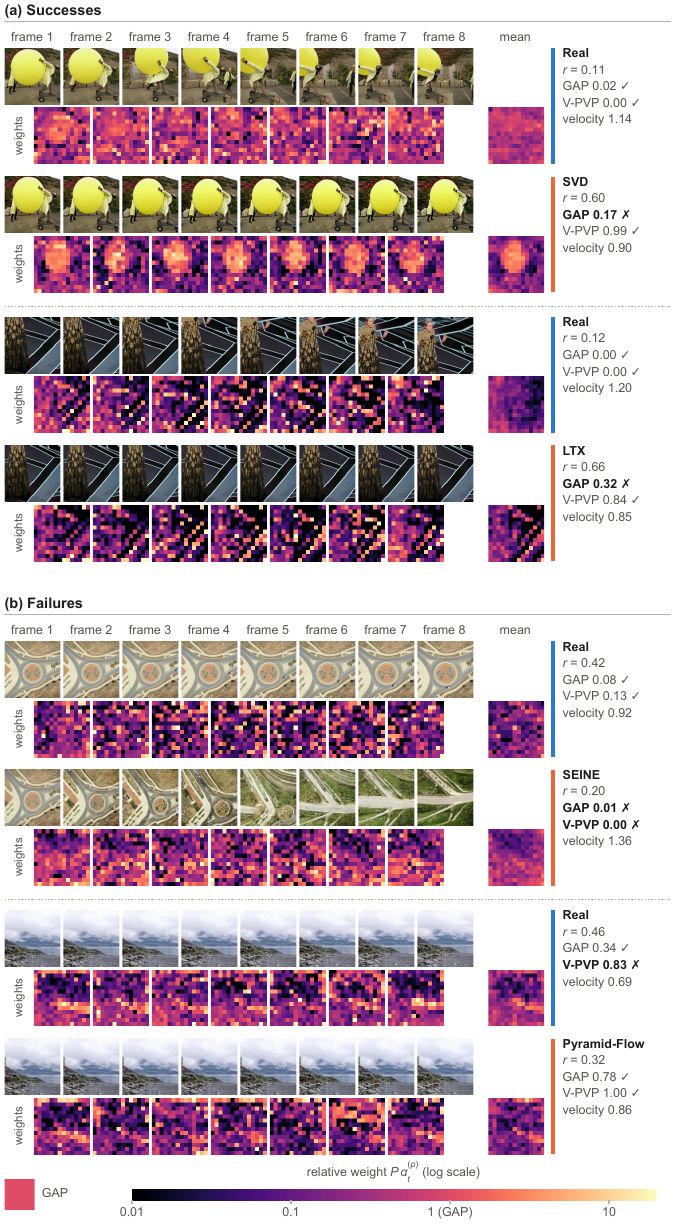}
    \caption{\textbf{Velocity gated weights on (a) successes and (b) failures of V-PVP.} Each scene pairs a real clip (blue) with an image-to-video fake generated from its first frame (orange): eight frames, the weights between consecutive frames (GAP $=1$), and their mean. Right: weight persistence $r$ (correlation of the weights between consecutive frame pairs), probability of being generated (wrong decisions in bold), and mean patch velocity relative to real test clips.}
    \label{fig:weight_maps}
\end{figure}

\begin{table}[h]
\centering
\caption{Sample-level error correlation between the two single-stream models. Mean across three random seeds with standard deviation.}
\label{tab:error_corr}
\begin{tabular}{lc}
\toprule
Quantity & Value \\
\midrule
$\phi(e_z, e_r)$ & $0.103 \pm 0.024$ \\
$P(e_r = 1)$ marginal & $0.182$ \\
$P(e_r = 1 \mid e_z = 1)$ conditional & $0.214$ \\
\bottomrule
\end{tabular}
\end{table}

The $\phi$ coefficient is well below the strong-correlation reference of 0.5, indicating that the errors of the two streams are weakly correlated. The conditional error rate $P(e_r = 1 \mid e_z = 1) = 0.214$ is moderately above the marginal rate $P(e_r = 1) = 0.182$, indicating residual but limited dependence rather than full statistical independence. This is consistent with two independent derivation paths corresponding to distinct but not strictly orthogonal discriminative evidence.

The second angle is on which sample subsets the $r$ stream is not replaceable by $\tilde{z}$. The residual term $C$ in Step 2 projects the inter-patch structure onto the weight deviation $\alpha^{(p)} - \omega^{(p)}$, so it is informative when that structure is itself discriminative. Section~\ref{sec:motivation} identifies the discriminative structure as the homogeneous, low-rank patch motion of generated videos, which appears as a large $\kappa_{\text{spec}}$ (Figure~\ref{fig:readout-bottleneck}(a)). On high-$\kappa_{\text{spec}}$ samples this Loss 1 evidence is pronounced and $C$ can carry it. On low-$\kappa_{\text{spec}}$ samples the inter-patch motion is heterogeneous and real-like, the Loss 1 evidence is weak, and the channel-level magnitude signal becomes the primary available evidence. If the two streams indeed carry the two types of evidence, the marginal gain of $r$ over $\tilde{z}$ should be larger on low-$\kappa_{\text{spec}}$ samples than on high-$\kappa_{\text{spec}}$ samples. We bucket the test videos into four equal-frequency groups Q1 to Q4 by $\kappa_{\text{spec}}$ from low to high, and compute
\begin{equation}
\Delta_q = \mathrm{AUC}(\tilde{z} + r)_q - \mathrm{AUC}(\tilde{z})_q
\end{equation}
within each bucket. The marginal gain is largest on Q1 and decreases toward Q4, matching the predicted division of labor that $\tilde{z}$ is sufficient when the Loss 1 evidence is strong and $r$ becomes necessary when it is weak. Combined with the weak error correlation in Table~\ref{tab:error_corr}, this provides finer evidence for the differentiated contributions of the two streams than the overall 1.27 AUC gain alone.
\section{Visualization of Velocity-Gated Weights}
\label{app:weights}

Figure~\ref{fig:weight_maps} shows the relative weight $P\,\alpha_t^{(p)}$ that V-PVP assigns to each patch: $1$ is the uniform GAP weight, and values above or below $1$ mark patches that V-PVP emphasizes or suppresses. Each real clip is paired with an image-to-video fake generated from its first frame, so that content is matched and only motion differs. In the successes~(a), the high-weight patches of the generated clips stay on the same region in every frame pair, such as the balloon that the fake keeps almost in place, whereas those of the real clips are spread over several regions and shift with the motion. This is consistent with Section~\ref{sec:motivation}: real videos carry richer, more heterogeneous inter patch motion than generated ones. GAP gives every patch the same weight, averages this difference away, and scores both fakes as real, while V-PVP classifies all four clips correctly. The failures~(b) arise when this cue is ambiguous: a fake with coherent, camera-like global motion has shifting weights like a real clip and is accepted, while a nearly static real clip has persistent weights like a generated one and is rejected. All cases hold in three training seeds; weights and scores are from one.

\section{Broader Impact}
\label{app:impact}

The development of AI-generated video detection methods such as V-PVP has
important societal and ethical implications. Our work supports video authenticity
verification as synthetic videos become increasingly realistic and easier to
distribute. Such detectors can assist journalistic provenance checks,
platform-side moderation, forensic triage, and auditing of emerging video
generation models. The lightweight readout, with only about 0.5M trainable
parameters, also makes recalibration more accessible to smaller research groups,
public-interest organizations, and non-frontier laboratories.

At the same time, AI-generated video detectors should not be treated as definitive
proof of authenticity or fabrication. False positives may unfairly cast doubt on
genuine videos, while false negatives may create misplaced confidence in
synthetic or manipulated content. Public release of detection tools may also
indirectly inform attempts to evade future detectors. We therefore view V-PVP as
a decision-support tool rather than an autonomous arbiter of authenticity. In
practical deployments, its outputs should be combined with complementary
provenance signals, such as metadata, watermarking, cryptographic provenance, or
human review, especially in high-stakes settings.
\section{Future Works}
\label{app:limitations}

While V-PVP shows strong cross-generator performance on AIGVDBench and GenVidBench-143k, its current evaluation is still bounded by the available benchmark protocols. These benchmarks are useful for testing whether a detector transfers to unseen generators, but they cannot fully cover the diverse conditions under which videos circulate online. In practice, a video may be recompressed, resized, cropped, screen-recorded, edited, captioned, or re-uploaded multiple times before being inspected, and these propagation steps may interact with temporal cues in ways that are not completely captured by existing datasets. Our implementation also follows a simple and fair clip-sampling protocol, using uniformly sampled clips of fixed temporal length. This choice makes comparisons controlled, but videos with sparse artifacts, abrupt scene cuts, or uneven motion may benefit from aggregating evidence across multiple temporal windows. Another open direction is adaptation beyond the frozen-backbone setting. Keeping the backbone fixed is central to our analysis of the readout bottleneck and gives V-PVP its parameter efficiency, yet domain-specific recalibration or lightweight adapters may further improve performance when the target videos differ substantially from common video pretraining corpora.

\textbf{Future Works.} Future work can extend V-PVP along several practical directions. A natural next step is to evaluate the method under realistic video propagation pipelines, including compression, resizing, frame-rate conversion, cropping, screen recording, and multi-stage social-media re-upload. Such evaluation would better connect benchmark performance with deployment conditions. Another promising direction is adaptive temporal evidence aggregation. Instead of relying on a single uniformly sampled clip, a detector could combine predictions from multiple temporal windows, use scene-boundary-aware sampling, or estimate uncertainty when the available temporal evidence is limited. V-PVP may also benefit from lightweight domain adaptation, such as generator-specific recalibration, LoRA-style adapters, or test-time adaptation, while retaining most of the efficiency advantage of the frozen-backbone design. More broadly, our results suggest that AIGV detectors should avoid collapsing patch-level temporal evidence too early. Extending this principle to larger video foundation models, multimodal video-language encoders, CNN-based backbones, and hybrid architectures may lead to more general and deployment-ready detection frameworks.

\end{document}